\documentclass[journal]{IEEEtran}
\usepackage{graphicx}
\usepackage{epstopdf}
\usepackage{cite}
\usepackage{algorithm}
\usepackage{booktabs}
\usepackage{multirow}
\usepackage{array}
\usepackage{subfigure}
\usepackage{algpseudocode}
\usepackage{amsmath}
\usepackage{graphics}
\usepackage{epsfig}
\usepackage{color}
\usepackage{bm}
\usepackage{amsmath}
\usepackage{multirow}
\usepackage{graphicx}
\usepackage{array}
\usepackage{subfigure}
\usepackage{algpseudocode}
\usepackage{amsmath}
\usepackage{graphics}
\usepackage{epsfig}
\usepackage{booktabs}

\usepackage[colorlinks,
            linkcolor=blue,
            anchorcolor=black,
            citecolor=black]{hyperref}
\usepackage{color}
\usepackage{bm}
\usepackage{amsfonts}
\usepackage{amsmath}
\usepackage{hyperref}
\usepackage{cite}

\ifCLASSINFOpdf

\else

\fi
\hypersetup{hidelinks,
	colorlinks=true,
	allcolors=blue,
	pdfstartview=Fit,
	breaklinks=true}
\ifCLASSINFOpdf
\else
\fi
\begin{document}
%

\title{
Bilateral Personalized Dialogue Generation with Contrastive Learning 
}
%
%
%

\author{Bin~Li,
        Hanjun Deng
\thanks{
\par Bin Li is with College of Electrical and Information Engineering, and with the Key Laboratory of Visual Perception and Artificial Intelligence of Hunan Province, Hunan University, Changsha, 410082, China.
(libincn@hnu.edu.cn).}}
%

%
%

\markboth{Journal of \LaTeX\ Class Files,~Vol.~14, No.~8, August~2015}%
{Shell \MakeLowercase{\textit{et al.}}: Bare Demo of IEEEtran.cls for IEEE Journals}
%



\maketitle

\begin{abstract}
Generating personalized responses is one of the
major challenges in natural human-robot interaction. Current
researches in this field mainly focus on generating responses
consistent with the robot's pre-assigned persona, while ignoring the user's persona. Such responses may be inappropriate or even offensive, which may lead to the bad user experience.
Therefore, we propose a Bilateral Personalized Dialogue Generation (BPDG) method for dyadic conversation, which integrates user and robot personas into dialogue generation via designing a dynamic persona-aware fusion method. 
To bridge the gap between the learning objective function and evaluation metrics, the Conditional Mutual Information Maximum (CMIM) criterion is adopted with contrastive learning to select the proper response from the generated candidates. Moreover, a bilateral persona accuracy metric is designed to measure the degree of bilateral personalization.  Experimental results demonstrate that, compared with several state-of-the-art methods, the final results of the proposed method are more personalized and consistent with bilateral personas in terms of both automatic and manual evaluations.
\end{abstract}


%
\IEEEpeerreviewmaketitle

\section{Introduction}
%
%
%
%

\IEEEPARstart{O}{ne} of the major challenges in human-robot interaction  is to develop an intelligent agent to generate natural, personalized, information-rich, and consistent responses\cite{adiwardana2020towards}. For this purpose, the dialogue agents have to learn to express personalized information appropriately like humans. Currently, personalized dialogue agents have been widely applied in various human-robot interaction scenarios, such as intelligent personal assistants\cite{de2020intelligent}, public service robots\cite{sheth2019cognitive}, wearable devices\cite{queiros2018smartwalk}, etc. The agents with personalization are considered reliable and trustworthy, and can gain the user's confidence and trust\cite{2}.
\par In the past decades, personalization has played an important role in the dialogue system and attracted wide attention\cite{3,4,Songhaoyu,Qian,Zheng}. According to the different ways of personalized information modeling, the existing personalized dialogue systems are mainly categorized into two types: implicit personalization\cite{4,Songhaoyu} and explicit personalization\cite{Qian,Zheng}.
\begin{figure}
	\centering{
		{\includegraphics[width=7.5cm]{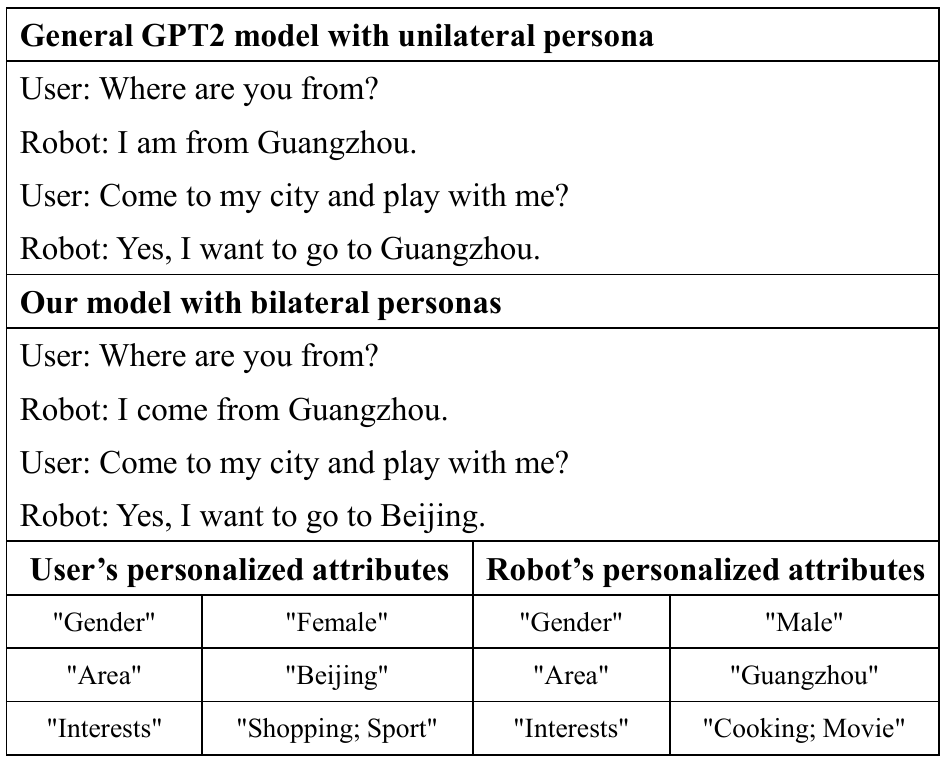}}
	}
	\caption{Exemplar dialogues with/without bilateral persona-consistent in dyadic conversation. The general GPT2 model with unilateral persona can generate a response that only meets the robot's persona, but ignores the persona of the other party. The proposed method can incorporate bilateral personas and generate a response that matches the personas of both parties.}
	\label{fig1}
\end{figure}
\par The implicit personalized dialogue system models personas with unstructured natural language utterances (e.g., ``I am a musician.'', ``I like to play the guitar.''), where the persona is implicitly mapped into the hidden state vectors. However, these implicit space mapping methods are poor in interpretability and may be over-fitting during the training process. Besides, the given utterances are mostly short and limited to only a few personas, the model may fail to utilize the persona properly when generating responses\cite{xu2020neural}. Indeed, implicit personalized dialogue corpus \cite{4} reflecting personas in every response is also different from the way of interpersonal conversation. 
\par The explicit personalized dialogue system models the personas with the structured personalized attributes, which are explicitly formatted as different key-value pairs (e.g.,
$<$Gender, Female$>$, $<$Area, Beijing$>$).
Such explicit persona modeling is more straightforward and interpretable. Specifically, the explicit personalized dialogue corpora\cite{Qian,Zheng} are crawled on a large scale from social networking sites, such as Weibo\footnote{https://www.weibo.com}, where people may unintentionally show their personality during the conversation. However, the explicit personalization in \cite{Qian}  models the robot's persona in the form of a pre-assigned profile and only emphasizes unilateral persona consisitency. The latest works \cite{Zheng,ZhengAAAI} incorporate the structured speaker's profile into the generated response to ensure the persona consistency of the speaker. Although these methods solve the problem of unilateral persona consistency to some extent, the robot may ignore the user's persona during the conversation. As a result, the generated responses may conflict with the user's personalized information. 
\par In the dyadic interpersonal conversations, both two interacting parties know each other's personalized information\cite{11}. When responding, the speaker should not only focus on their own personalized expression, but also consider the questions and persona of the other party\cite{12}. As is shown in Fig. \ref{fig1}, during the conversation, the robot should generate responses consistent with the robot's own personalized attributes (i.e., unilateral persona-consistent). Furthermore, the robot also  should know the user's persona and generates responses consistent with the user's personalized attributes (i.e., bilateral persona-consistent). Once these factors are ignored, it may annoy the user and reduce the user experience.

\par To solve the above problem, we propose a bilateral personalized dialogue generation (BPDG) method to generate responses consistent with both personas. Specifically, the BPDG method is based on the structure of language model with multi-task transfer learning. The proposed method optimizes three tasks simultaneously: the language model task, the persona presence prediction task, and the dialogue generation task. In the language model task, the dialogue utterances embedded with the corresponded personalized attributes and relative position are used to train the encoder. 
In the persona presence prediction task, the dialogue contextual encoding is used to predict the possibilities of the personas' presence in the response. More precisely, the encodings of the dialogue context, bilateral personas and the right-shifted outputs are fused with four different fusion methods (i.e., add, max, mean and dynamic) to demonstrate the importance of bilateral personas. In the dialogue generation task, the fused encoding is input into the decoder to generate response candidates with the diverse beam search strategy \cite{vijayakumar2016diverse}. Finally, in order to ensure the generated responses are more personalized and bilateral persona-consistent, we adopt the Conditional Mutual Information Maximum (CMIM) criterion with contrastive learning to select the final response from the diversified generated candidates. Thus, the proposed BPDG method can utilize bilateral personalized information to generate personalized and bilateral persona-consistent responses for better user experience in the human-robot interaction.
\par The main contributions of this article can be summarized as follows.
\begin{enumerate}
	\item We propose a novel BPDG method, which integrates the bilateral personas to generate responses consistent with both personas. To the best of our knowledge, this is the very first to propose the bilateral persona consistency in the personalized dialogue generation.
	\item A dynamic persona-aware fusion module is developed
	to adaptively control the encodings of the bilateral
	personalized information, the dialogue context, and the
	shifted right outputs for decoding to generate bilateral
	persona-consistent responses.
	\item We adopt the criterion of the CMIM with contrasti-ve learning, which briges the gap between the learning objective and evaluation metrics.
	\item Both automatic and manual evaluations show that our method outperforms state-of-the-art methods.
\end{enumerate}
\par The remainder of this article is structured as follows. Section II reviews the work related to the personalized dialogue system. Section III formulates the problem and details the proposed BPDG method.  Section IV fully describes the experimental setups. Automatic and human evaluations are illustrated and analyzed in detail in Section V and Section VI respectively. Finally, the conclusions and some possible future work are pointed out in Section VII.




\section{Related Work}



\subsection{Personalized Dialogue Generation}
Inspiring by the ``Big Five" \cite{14} in psychology, Mairesse  \emph{et al.} \cite{15} take the lead in incorporating the personalities into the framework of dialogue generation, thereby generating responses with recognizable personality. However, the personality of the ``Big Five" is extremely implicit and subtle. It is necessary to build rules to capture personality characteristics. Besides, it is a challenge to construct a corpus with limited and laborious collection. With the popularity of deep learning, hand-craft rule modeling is gradually replaced by data-driven modeling. Li \emph{et al.}\cite{16} first propose a personalized dialogue generation model, mapping the persona in natural utterance into distributed representation vectors on the seq2seq framework, which is benefited from the neural machine translation\cite{17}. Subsequently, there are other different methods used for personalized dialogue generation modeling, for example, Song \emph{et al.} \cite{Songhaoyu} adopt the CVAE method implicitly learns the responses that contain personalized information to generate personalized responses. Madotto \emph{et al.} \cite{19} design a personalized dialogue generation model with meta-learning. Yang \emph{et al.} \cite{20} describe an empirical survey of personalized dialogue generation via reinforcement learning.
The above method is effective, but it also faces the problem of generating general or bilateral-inconsistent responses. Different from the previous work, the proposed BPDG method further integrates personalized information from both parties into the pre-trained decoder-only framework, to generate bilateral persona-consistent responses with multi-task learning and transfer learning.
\begin{figure*}
	\centering{
		{\includegraphics[width=15.9cm]{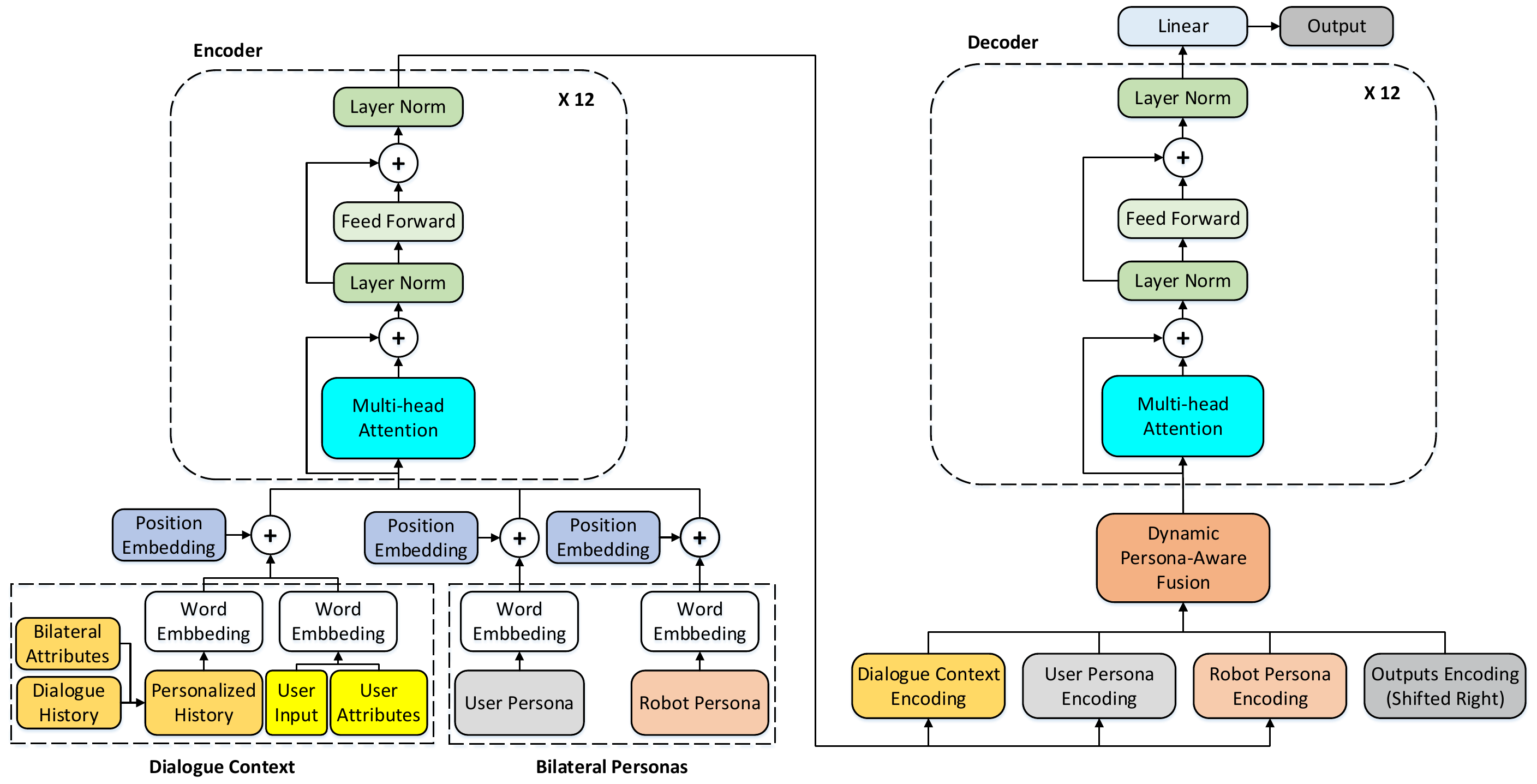}}
	}
	\caption{The overview of the proposed BPDG method. 
	}
	\label{fig2}
\end{figure*}
\subsection{Multi-task Transfer Learning}
Multi-task Transfer learning aims to extract and transfer the knowledge from the source domain to the target domain \cite{22} with different well-designed learning tasks, which has been very popular in the field of the NLP in the past decade\cite{21}. Recent advances in natural language generation rely on pre-training a large generative language model with a large corpus of unsupervised data. It mainly follows the two-stage paradigm of pre-training and fine-tuning. In the field of personalized dialogue generation, Zhang \emph{et al.} \cite{23} first introduce transfer learning into the two-stage personalized dialogue generation.
Wolf \emph{et al.} \cite{25} design a pre-trained dialogue generation model that jointly learns two tasks (e.g., next sentence prediction and language model) when finetuning. The experimental results show that multi-task learning can greatly improve the scores in automatic metrics. Golovanov \emph{et al.} \cite{28} integrate multi-task learning into the transfered model with shared parameters, and designs three sub-tasks, including language model task, dialogue generation task and expected risk task. These tasks are proven to improve the performance in human evaluation. Zheng \emph{et al.} \cite{ZhengAAAI} leverage target persona information in generating unilateral persona-consistent responses by designing three different tasks, including the language model, persona routing, and dialogue generation. In this article, apart from the language model task and dialogue generation task, we further design a persona prediction task for the dynamic persona-aware fusion module, adaptively fusing the encodings of different information for decoding, to generate responses consistent with bilateral personas.
\subsection{Contrastive Learning}
Contrastive Learning \cite{hadsell2006dimensionality, gutmann2012noise,chen2020simple} uses self-supervised methods to learn the representation of the  positive examples and negative examples. The contrastive learning method learns the general features of the corpus without labels by teaching the model which data is similar or different. In the field of natural language processing, contrasted learning has good performance in tasks such as language model task\cite{baltescu2015pragmatic}, image captioning\cite{dai2017contrastive}, text summarization\cite{liu2021simcls}. In the field of human-robot interaction, contrastive learning is conducive to capturing the information implicit in the dialogue\cite{cai2020group} and it is useful for filling the gap between learning objective function and evaluation metrics\cite{liu2021simcls}. Therefore, this paper introduces the conditional mutual information criterion in the bilateral personalized dialogue generation. By ranking the candidate responses through comparative learning, the final outputs can be rich in bilateral  personalized information.
\section{Proposed Method}
In the dyadic interpersonal conversation, both interacting parties have their own personas such as gender, area, individual interests, etc. Such information may be presented in the response. 
In the human-robot dialogue, given the user persona \emph{U}, the robot persona \emph{R}, the personalized history \emph{H} and the user input \emph{X}, the robot generates a natural, fluent and personalized response \emph{Y}, which can be formulated as follows:
\begin{equation}
Y=\underset{Y^{\prime}}{\arg \max } P\left(Y^{\prime} \mid X, H, U, R\right)
\label{euqtion001}
\end{equation}
where the user persona $U$ and the robot persona $R$ can be represented with the personal profile, which is formatted as a set of attributes composed of key-value pairs. Each attribute in the user persona $U=\left\{u_{1}, u_{2}, \ldots, u_{m}\right\}$ is a key-value pair $u_{i}=\left\langle k_{i}, v_{i}\right\rangle$. The robot persona $R$ is represented likewise. The personalized history is represented as $H=\left\{\left\{X_{1}^{U}, U\right\},\left\{X_{2}^{R}, R\right\}, \ldots,\left\{X_{l}^{R}, R\right\}\right\}$, where the superscript indicates the speaker, and the subscript indicates the number of the dialogue rounds. Each sentence is associated with the persona of the corresponded speaker. The user input $X=\left\{{X_{l+1}^{U}, U}\right\}$ contains the user current input $X_{l+1}^{U}$ with the user persona $U$.
\par Combining the user input $X$ and the personalized history $H$ into the context of the dialogue $C$, the equation (\ref{euqtion001}) can be further written as the equation (\ref{equation2}):
\begin{equation}
Y=\underset{Y^{\prime}}{\arg \max } P\left(Y^{\prime} \mid C, U, R\right)
\label{equation2}
\end{equation}
where the dialogue context $C = <H,X>$ represents that the personalized history $H$ is concatenated with the current user input $X$.
\par Fig. \ref{fig2}. is the overview of the proposed BPDG method. The BPDG method consists of the encoder, the dynamic persona-aware fusion module, and the decoder. Following the GPT2 framework, the encoder and decoder share the same weights and act as a backbone to learn the sentence representation. The encoder trains the language model with the dialogue context embedding and encodes the embedding of the user persona and the robot persona independently. The persona-aware fusion module is used for fusing the dialogue context encoding, the bilateral persona encodings and the shifted right outputs encoding. Afterwards, the fused encoding is sent into the decoder for generating several candidate responses with the diverse beam search strategy. Finally, the CMIM criterion is adopted to output a personalized and bilateral persona-consistent response.
\begin{figure}[]
	\centering{
		{\includegraphics[width=8cm]{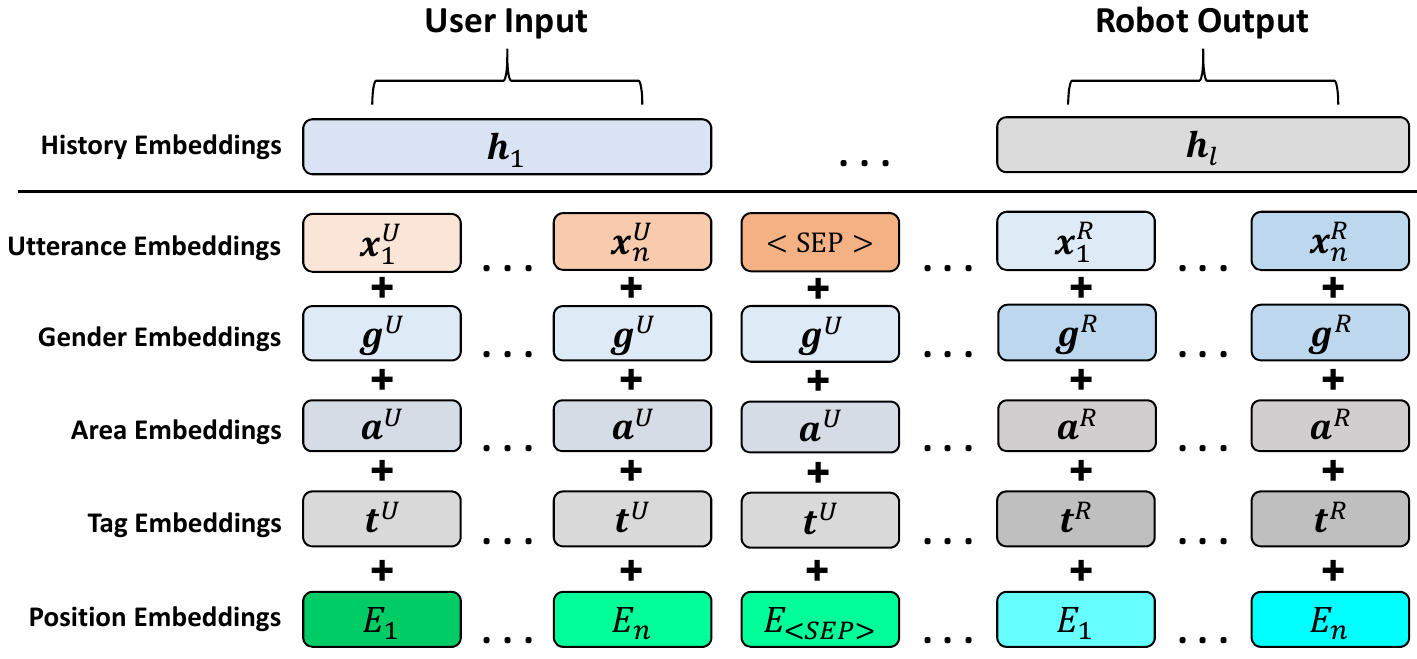}}
	}
	\caption{The structure of personalized history embeddings.
	}
	\label{fig3}
\end{figure} 
\subsection{Dialogue Context Modeling}
Dialogue context modeling means that each dialogue utterance embedding is added with the corresponded persona embedding and relative position embedding to obtain the embeddings of personalized history. The dialogue context embedding can be obtained by concatenating the embeddings of the personalized history and the current user input. The dialogue context encoding is obtained with the dialogue context embedding being encoded. The process can be described as follows:
\subsubsection{Utterence Embedding} The utterances of the user and the robot are first embedded with word embedding respectively. The $X_U $ represents the embedded user input, and the $ X_R $ represents the embedded robot output. Both embeddings are specified with the same length $n$. If the corresponded length does not reach the specified length, we use $<PAD>$ as a placeholder. Otherwise, a truncation operation is taken. The word embedding process is shown as follows:
\begin{equation}
\boldmath
X_{U}={\left\{{\bm{x}_{1}^{U}}, {\bm{x}_{2}^{U}}, {\bm{x}_{3}^{U}}, \ldots, {\bm {x}_{n}^{U}}\right\}}
\unboldmath
\end{equation}
\begin{equation}
\boldmath
X_{R}={\left\{{\bm{x}_{1}^{R}}, {\bm{x}_{2}^{R}}, {\bm{x}_{3}^{R}}, \ldots, {\bm {x}_{n}^{R}}\right\}}
\unboldmath
\end{equation}
where the $X_{U}$ is the embedding of the user input, the {{${\bm{x}_{i}^{U}}$}} is the word embedding of the \textit{i-th} token in the sentence input by the user, and the $X_{R}$ is the embedding of the robot response, the
{${\bm{x}_{i}^{R}}$}
is the word embedding of the \textit{i-th} token in the sentence output by the robot.
\subsubsection{Persona Embedding} Persona embedding means the utterances embedded with the corresponded personas attributes. As is mentioned before, the profile consists of three attributes: gender, area, and individual interests. The value of the gender is binary (i.e., 0 for male and 1 for female). The value of the area is represented with the index of the corresponded item in the look-up table. The items of the look-up table are sorted by the occurrence frequency of the area in the corpus. The individual interests are represented in a similar way. To take the operation of the user as an example, the process is shown in the equation (\ref{7}):
\begin{equation}
\begin{aligned}[c]
G_{U}&= \left\{\boldsymbol{g}_{1}^{U}, \boldsymbol{g}_{2}^{U}, \ldots, \boldsymbol{g}_{j}^{U}, \ldots, \boldsymbol{g}_{n}^{U}\mid \boldsymbol{g}_{j}^{U} = \boldsymbol{g}^{U} \right\} \\
A_{U}&= \left\{\boldsymbol{a}_{1}^{U}, \boldsymbol{a}_{2}^{U}, \ldots,
\boldsymbol{a}_{j}^{U}, \ldots, \boldsymbol{a}_{n}^{U} \mid \boldsymbol{a}_{j}^{U} = \boldsymbol{a}^{U} \right\} \\
T_{U}&=  \left\{\boldsymbol{t}_{1}^{U}, \boldsymbol{t}_{2}^{U}, \ldots, \boldsymbol{t}_{j}^{U}, \ldots, \boldsymbol{t}_{n}^{U} \mid \boldsymbol{t}_{j}^{U} = \boldsymbol{t}^{U} \right\} \\
\end{aligned}
\label{7}
\end{equation}
where the $\boldsymbol{g}^{U}$ represents the word embedding of the user's gender extracted from the profile, the $\boldsymbol{g}_{j}^{U}$ represents the gender embedding $\boldsymbol{g}^{U}$ corresponding to the position \emph{j} in the user input embedding $X_{U}$, $j \in[1, n]$. The $\boldsymbol{a}^{U}$ and $\boldsymbol{t}^{U}$ represent the word embedding of the user's area and individual interests tag extracted from the profile respectively. For multiple individual interests, we take the average of the first-three embeddings of individual interests.
\par The relative position embedding \cite{31} is adopted to make the embedded tokens more sensitive to the position in the sentence for further attention operation. The position embedding is written as follows:
\begin{equation}
\begin{aligned}
E_{{i}}(2k) &=\sin \left(\frac{{i}}{10000^{\frac{2 k}{{d_{model}}}}}\right) \\
E_{{i}}(2k+1) &=\cos \left(\frac{{i}}{10000^{\frac{2 k}{{d_{model}}}}}\right)
\end{aligned}
\label{position}
\end{equation}
where $i$ is the position of the token in the sentence, $k$ represents the \textit{k-th} dimension of the word embedding, ${d_{model}}$ is the fixed embedding dimension.
\subsubsection{Personalized history embbedings}
The Fig. \ref{fig3}. shows the structure of personalized history embeddings. The personalized history embeddings are a combination of the aforementioned three types of embeddings, i.e., the embeddings of the utterance, the persona embeddings, and the position embeddings, with the $<SEP>$ being used as the separator.
Specifically, the personalized history embeddings are formatted utterance by utterance with concatenation, which can be wrriten as the equation (\ref{phd}).
\begin{equation}
\begin{aligned}
\boldsymbol H &=\text {Concat} \left\{\boldsymbol h_{1}, \boldsymbol h_{2}, \ldots,\boldsymbol h_{j}, \ldots, \boldsymbol h_{l}\right\} \\
\boldsymbol h_{j}&=\begin{cases}
D_{U}, \text { if }\boldsymbol{\bmod} (j, 2)=0\\
D_{R}, \text { if }\boldsymbol{\bmod} (j, 2)=1
\end{cases},j \in[1, l]
\end{aligned}
\label{phd}
\end{equation}
where the Concat $\left\{\right\}$ represents the operation of concatenation,  $l$  represents the total number of rounds of the personalized history,
$h_{j}$ represents the personalized history of the $j$ round, $j \in[1, l]$. For each utterance, the personalized history embeddings are calculated via aligning the embeddings by token and performing token-wise aggregation. This process can be expressed as follows:
%
\begin{equation}
D_{U}=\text {Add}\left(X_{U}, G_{U}, A_{U}, T_{U}, E\right)
\end{equation}
\begin{equation}
D_{R}=\text {Add}\left(X_{R}, G_{R}, A_{R}, T_{R}, E\right)
\end{equation}
where the Add $\left(\right)$ represents the token-wise addition operation of the different embeddings with the same embedded length.
\subsubsection{Dialogue Context embedding} The personalized history embeddings and the user current input at the $l+1$ round are concatenated into the dialogue context embedding $\boldsymbol C$, which can be expressed as follows:
%
%
\begin{equation}
\boldsymbol C =\text {Concat} \left\{\boldsymbol H, \boldsymbol h_{l+1}\right\} 
\end{equation}
\par Finally, the dialogue context encoding $E_{C}$ is obtained after the dialogue context embedding $\boldsymbol C$ is encoded.
\subsection{Bilateral Profile Modeling}
To take advantage of the bilateral personas in the dialogue generation, the explicit form of persona, i.e., the profile, is used in the proposed method. Word embedding is performed on the profile text to represent the semantic information in the same way as the utterance, which will benefit the further processing.
Specifically, the word embedding of the user persona $\boldsymbol{U}$ and the robot persona $\boldsymbol{R}$ can be written as follows:
\begin{equation}
\boldsymbol{U}=\left\{\boldsymbol{u}_{1}, \boldsymbol{u}_{2}, \boldsymbol{u}_{3} \mid \boldsymbol{u}_{i}=\{\boldsymbol{s}, \boldsymbol{v}\}, i=1,2,3\right\}
\label{equation3}
\end{equation}
\begin{equation}
\boldmath
\boldsymbol{R}=\left\{\boldsymbol {r}_{1}, \boldsymbol{r}_{2}, \boldsymbol{r}_{3} \mid \boldsymbol{r}_{i}=\left\{\boldsymbol{s}^{\prime}, \boldsymbol{v}^{\prime}\right\}, i=1,2,3\right\}
\unboldmath
\label{equation4}
\end{equation}
where each attribute $\boldsymbol u_{i}$ in the embedded user persona $\boldsymbol{U}$ is the word embedding of the key-value pair. The embedded user persona $\boldsymbol{U}$ is the concatenation of the three attributes corresponding to gender, area, and individual interests respectively. The comma is used as the separator to concatenate each key-value pair. The  embedded robot persona $\boldsymbol{R}$ is formatted likewise.
\par Further, the embedded user persona $\boldsymbol{U}$ with relative position embedding $E$ is input into the encoder to obtain the user persona encoding $E_{U}$, while the embedded robot persona $\boldsymbol{R}$ turns into the $E_{R}$ is in the same way. The above process is implemented independently, which means that the $E_{U}$ and $E_{R}$ do not participate in the training of the encoder. 
\begin{figure}[]
	\centering{
		{\includegraphics[width=7.5cm]{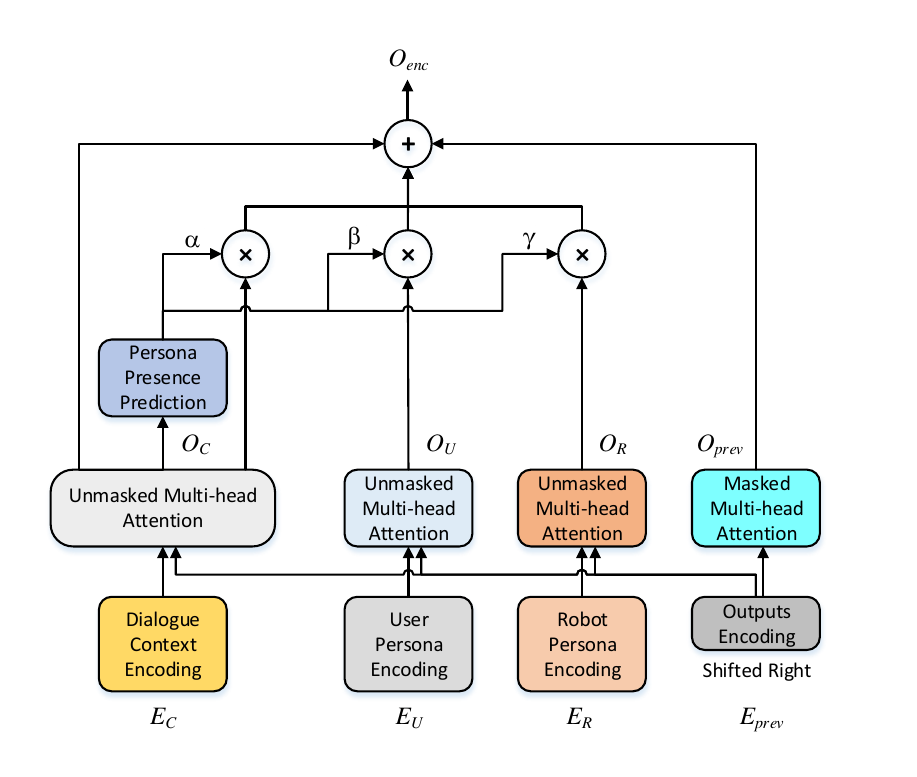}}		
	}
	\caption{The structure of the dynamic persona-aware fusion module.}
	\label{fig4}
\end{figure}
\subsection{Persona-aware Fusion Module}
In the bilateral personalized dialogue generation, two critical problems have to be addressed for appropriate persona expression: (1) when to express persona, and (2) whose persona should be expressed. Therefore, we propose dynamic persona-aware fusion to predict the presence of the bilateral personas and adaptively fuse them into the encodings for the further personalized response generation.
Fig. \ref{fig4}. shows the structure of the dynamic persona-aware fusion module. The persona-aware means that the presence of the persona in the generated response can be predicted with the dialogue contextual encoding $O_{C}$ obtained from the attention operation. The prediction probability is used to dynamically weighted to the corresponded attention encoding for fusion.
\subsubsection{Encoding Attention Mechanism} In order to effectively utilize the information of the encodings, we design different encoding attention mechanisms. Each encoding from the encoder participates in the unmasked multi-head attention mechanism. The masked multi-head attention mechanism is designed to avoid feeding the shifted-right ground-truth tokens when training. 
The $ prev $ represents the previously decoded output word, which turns into the outputs encoding $E_{prev}$ with word embedding and position embedding. The $E_U$ is input into the unmasked multi-head attention network to obtain the user personalized encoding $O_U$ and the robot personalized encoding $O_{R}$ is obtained in the same way. The unmasked multi-head attention process is shown as follows:
\begin{equation}
O_{U}=\text {Multi-head }\left(E_{\emph{prev}}, E_{U}, E_{U}\right)
\end{equation}
\begin{equation}
O_{R}=\text {Multi-head }\left(E_{\emph{prev}}, E_{R}, E_{R}\right)
\end{equation}
where the $E_{prev}$ is the query, the $E_{U}$ is both the key and the value in the unmasked multi-head mechanism, the operation of the robot personalized encoding $O_R$ is the same.
\par The context encoding $E_C$ and the outputs encoding at the previous moment $E_{\emph{prev}}$ are used to obtain the dialogue contextual encoding $O_C$ with the unmasked multi-head attention network, as is shown in equation (\ref{contextual}):
\begin{equation}
O_{C}=\text {Multi-head}\left(E_{\emph{prev}}, E_{C}, E_{C}\right)
\label{contextual}
\end{equation}
where the $E_{prev}$ is the query, the $E_{C}$ is both the key and the value in the unmasked multi-head mechanism.
\par Furthermore, a masked multi-head attention network is used to obtain the previous outputs encodings $O_{prev}$, as is shown in equation(\ref{maskmultihead}):
\begin{equation}
O_{prev}=\text  {MaskedMulti-Head}\left(E_{\emph {prev}}, E_{\emph {prev}}, E_{\emph {prev}}\right)
\label{maskmultihead}
\end{equation}
where the $E_{prev}$ is the query, the key, and the value in the masked multi-head mechanism.
\subsubsection{Persona Presence Prediction}
The presence of the bilateral personas in the response is predicted for the dynamic persona-aware fusion of different encodings. To train a subnetwork for this task, we construct a heuristic script to label the utterance with three labels based on the presence of bilateral personas. The dialogue contextual encoding $O_{C}$ is used to predict the probability of three types of information, which is presented in the response sentence. The loss function is designed as follows:
\begin{equation}
L_{P}(\theta)=-\sum_{j=1}^{3} l_{j} \log P_{\theta}\left(l=j \mid O_{C}\right)
\label{fusion_task}
\end{equation}
where the $l_{j}$ represents the label of different persona type, $ \log P_{\theta}\left(l=j \mid O_{C}\right)$ represents the probability of the persona type predicted in the generated response based on the dialogue contextual encoding $O_{C}$.
\subsubsection{Persona Encoding Fusion} To utilize personalized information of different encodings, the dynamic encoding fusion is designed to adaptively control the persona presented in the generated response. The probability of three categories, which is used as the persona-aware weight for dynamic encoding fusion.  Specifically, each category is operated with the \emph{softmax} operation, which can be shown in the equation (\ref{probo}):
\begin{equation}
P_{\theta}\left(l=j \mid O_{C}\right)=\frac{\exp{O}_{C}^{(j)}}{\sum_{i=3} \exp{O}_{C}^{(i)}}, j=0,1,2
\label{probo}
\end{equation}
where the ${O}_{C}^{(j)}$ represents the dialogue contextual encoding $O_{C}$ corresponding to the \emph{j-th} label, which is obtained with a two-layer perception network with global and average pooling.
\par Each prediction probability is defined as the persona-aware weight as follows:
\begin{equation}
\alpha=P_{\theta}\left(l=0 \mid O_{C}\right)
\end{equation}
\begin{equation}
\beta=P_{\theta}\left(l=1 \mid O_{C}\right)
\end{equation}
\begin{equation}
\gamma=P_{\theta}\left(l=2 \mid O_{C}\right)
\end{equation}
where the $\alpha$ represents the probability of the user personalized information presented in the response, the $\beta$ represents the probability of the robot personalized information presented in the response, and the $\gamma$ represents the probability that personalized information does not present in the response, which means the context-related.
Three different encodings are dynamically weighted and fused, with the dialogue contextual encoding $O_{C}$ and the previous outputs encodings $O_{prev}$. These encodings together form the fused encoding $O_{enc}$, as is shown in equation (\ref{o_enc}):
\begin{equation}
O_{e n c}=\alpha O_{U}+\beta O_{R}+(\gamma+1) O_{C}+O_{p r e v}
\label{o_enc}
\end{equation}
where $\alpha + \beta + \gamma = 1$.
\par After fusing the different encodings with the dynamic persona-aware fusion module, the fused encoding $O_{enc}$ is input into the decoder for dialogue generation.
\subsection{Multi-task Learning for Dialogue Generation}
To train the proposed BPDG model, three different tasks have to be accomplished including language model task, persona prediction task and dialogue generation task. These tasks will be described below.
\subsubsection{Language Model Task}
A pre-trained model is first utilized to initialize the parameters of the GPT2 framework. In order to bridge the gap between the data utilized in the pre-training and fine-tuning stage, the language model is then adopted to finetune with the bilateral personalized dialogue dataset mentioned in Section IV-A. The language model is trained by optimizing the standard maximum log-likelihood loss, as shown in equation (\ref{language model}):
\begin{equation}
L_{L M}(\varphi)=-\sum_{i} \log P_{\varphi}\left({x}_{i} \mid {x}_{i-k}, \ldots, {x}_{i-1}\right)
\label{language model}
\end{equation}
where $\varphi$ represents the parameters of the language model, $k$ is the size of the context window, and ${ {x}_{i-k}, \ldots, {x}_{i-1}}$, ${x}_{i}$ is sequence of tokens sampled from the training corpus.
\subsubsection{Persona Prediction Task}
The persona prediction task is to predict the persona presence according to the contextual encoding $O_{C}$. The loss function is shown in the equation (\ref{fusion_task}). As a result, the prediction probability is used to dynamically weighted the different encodings to get the fused encoding $O_{enc}$. Finally, the $O_{enc}$ is input into the decoder for dialogue generation.
\subsubsection{Dialogue Generation Task}
The dialogue generation task is designed to to generate the bilateral personalized responses, the loss function of the dialogue generation task is shown as equation (\ref{dialogue_generation}):
\begin{equation}\small
\begin{aligned}
L_{D}(\varphi) &=-\sum_{i} \log P_{\varphi}\left({y}_{i} \mid {y}_{0}, \ldots, {y}_{i-1}, E_{C}, E_{U}, E_{R}\right) \\
&=-\sum_{i} \log P_{\varphi}\left({y}_{i} \mid O_{enc}\right)
\end{aligned}
\label{dialogue_generation}
\end{equation}
where $y_i$ represents the \textit{i-th} word generated by the decoder, and ${y_{0}, \ldots,y_{i-1}}$ is a sequence of previously generated words. Identically, the input of the decoder also can be written as the fused encoding.
\par Finally, the joint loss function of the entire model is presented in equation (\ref{all}):
\begin{equation}
\begin{aligned}
L(\varphi, \theta)=L_{D}(\varphi)+\lambda_{1} L_{L M}(\varphi)+\lambda_{2} L_{P}(\theta)
\end{aligned}
\label{all}
\end{equation}
where the $\lambda_{1}$ and $\lambda_{2}$ are the balance weights of the loss function of the language model task and the loss function of the persona prediction task respectively.
\subsection{Candidate Selection with CMIM}
After the dialogue generation via dynamic persona-aware fusion, the response is output with the decoding strategy.
However, the top-ranked candidates with the beam search strategy are usually general, short, or even unrelated \cite{kulikov2019multi}, so that responses related to both personas and history conditions often fail to achieve high ranking scores. To remedy this, the criterion of CMIM \cite{CMIM} is adopted to constrain the personalized and history information that reflects in the response. Specifically, the BPDG method utilizes the diverse beam search strategy to generate the best diversed top-20 candidate list, and adopts the CMIM criterion to select the response with the largest conditional mutual information value as the final response.
\begin{figure}[]
	\centering{
		{\includegraphics[width=6.9cm]{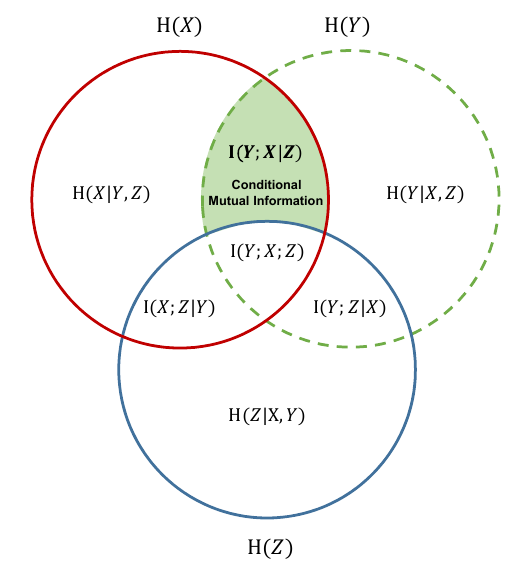}}
	}
	\caption{Illustration of conditional mutual information. The circles represent the information entropies of the different variables. The dashed circle represents the information entropy of the generated responses.
	 }
	\label{fig5}
\end{figure}
\subsubsection{Conditional Mutual Information Modeling}
In order to simplify the modeling process, the user persona $U$, the robot persona $R$, and personalized history information $H$ can be regarded as the condition $Z$. The illustration of conditional mutual information is shown in Fig. \ref{fig5}. Given the different conditions, i.e., $H,U,R$ in the same dialogue, the value of conditional mutual information $CMI_{v}$ of the user input $X$ and the robot generated candidate response $Y_{i}$ can be expressed as equation (\ref{CMI_V}):
\begin{equation}
\begin{aligned}
CMI_{v}(Y_{i}) &\equiv{\rm{I}(\mathit{Y_{i}} ; \mathit{X} \mid \mathit{H, U, R})} \\
&=\underbrace{\rm{H}(\mathit{Y_{i}} \mid \mathit{H, U, R})}_{\text { Relevance Ranking }}- \underbrace{\rm{H}(\mathit{Y_{i}} \mid \mathit{X, H, U, R})}_{\text { Dialogue Generation }} 
\end{aligned}
\label{CMI_V}
\end{equation}
where the CMIM criterion can be modeled with two terms, i.e., the dialogue generation item and the relevance ranking item.
\par According to the definition of the CMI \cite{CMIM}, the maximum of the equation (\ref{CMI_V}) can be achieved by solving the following optimization problem
\begin{equation}
Y^{*} = \arg\max_{Y_{i}} \log \frac{P(Y_{i} \mid X, Z)}{P(Y_{i} \mid Z)}
              \label{CMIV}
\end{equation}
where the $Y^{*}$ represents the final response in the top-20 candidate list. The $P(Y_{i}|X,Z)$ and $P(Y_{i}|Z)$ are corresponded to the dialogue generation term and relevance ranking term in equation (\ref{CMI_V}) respectively. 
\par The $P(Y_{i}|X,Z)$ is the probability of the generated response conditioned on the input and the context with the word granularity, while the $P(Y_{i}|Z)$ is the relevance of the response to the contextual content with the sentence granularity. Therefore, the $P(Y_{i}|X,Z)$ and $P(Y_{i}|Z)$ of equation (\ref{CMIV}) are not optimized jointly. 
\subsubsection{Dialogue Generation}
\par The $P(Y_{i}|X,Z)$ can be modeled with the BPDG model and calculated with the diverse beam search score. By substituting the $Z$ with $H,U,R$, the $P(Y_{i}|X,H,U,R)$ can be written as equation (\ref{BPDGmodeling}):
\begin{equation}
\small
\begin{aligned}
\log P(Y_{i} \mid X, H, U, R)& \equiv \log P_{\psi}\left(Y_{i} \mid X, E_{H}, E_{U}, E_{R}\right) \\
&= \log P_{\psi}\left(Y_{i} \mid O_{enc}\right) 
\end{aligned}
\label{BPDGmodeling}
\end{equation}
where the $\psi$ represents the parameters of the trained BPDG model, containing all the parameters in (\ref{all}).
\subsubsection{Relevance Ranking with Contrastive Learning}
After the candidate list is generated with the diverse beam search strategy, each candidate can be ranked with relevance ranking. Given the condition $Z$, i.e., the user persona $U$, the robot persona $R$, and the personalized history $H$, the relevance probability is calculated as
\begin{equation}
\begin{aligned}
\log P(Y_{i} \mid H, U, R) &= \frac{\log P_{\phi}( Y_{i}, H, U, R)}{\log P_{\phi}(H,U,R)} \ \ \ \\
&\propto \log P_{\phi}( Y_{i}, H, U, R)
\label{co_oc}
\end{aligned}
\end{equation}
where the $\phi$ represents the parameters of the content relevance classifier model trained on the corpus, the co-occurrence probability $P_{\phi}(H, U, R)$ is not related to $Y_i$ which can be omitted, and
the $P_{\phi}(Y_{i}, H, U, R)$ represents the co-occurrence probability of $Y_{i}$, $H$, $U$ and $R$ in the same dialogue.
\par Therefore, the relevance probability of each candidate can be modeled with the content relevance classifier $P_{\phi}(Y_i, H, U, R)$, we adopt the contrastive learning training method \cite{cai2020group} to perform relevance ranking step. To construct the training corpus for content relevance classifier, the $Y$, $H$, $U$, and $R$ from the corpus $\mathbb{D}$ are used as positive training samples, which marked as, while the $Y^{\prime}$, $H$, $U$, and $R$ from different corpus are sampled as negative samples, which is inspired by the practice in \cite{36}. The cross-entropy loss function used to train content relevance classifier $\phi$ is as follows
\begin{equation}
\small
\begin{aligned}
L_{\phi}\boldsymbol= &
-\frac{1}{N} \sum_{(Y, H, U, R) \in \mathbb{D}} \log  P\left((Y, H, U, R)^{+} ; \boldsymbol{\phi}\right) \\
&-\frac{1}{N} \sum_{(Y^{\prime}, H, U, R) \in \mathbb{D}} \log \left[1-P\left(Y^{\prime}, H, U, R)^{-} ;  \boldsymbol{\phi}\right)\right].
\end{aligned}
\end{equation}
%
\subsubsection{Candidate Selection} With the BPDG model and the content relevance classifier, the optimization problem in (\ref{CMIV}) can be written as folows.
\begin{equation}
Y^{*}=\arg\max_{Y_{i}}\log \frac{P_{\psi}\left(Y_{i} \mid O_{enc}\right)}{P_{\phi}(Y_{i}, H, U, R)}
\end{equation}
where $Y_{i}$ represents the response candidates.
\par
In order to further fintune the relevance of the generated response to the contextual content, an additional penalty parameter $\lambda_{3}$ is introduced for the relevance ranking item. The too-large value of $\lambda_{3}$ may reduce the relevance of the final response to the user input. Thus, the final response can be selected by equation (\ref{CMIM_V}):
\begin{equation}
\small
Y^{*} = \arg \max _{Y_{i}} \log P_{\psi}\left(Y_{i} \mid O_{enc}\right) - \lambda_{3} \log P_{\phi}\left(Y_{i}, H, U, R\right)
\label{CMIM_V}
\end{equation}
\section{Experiments}
\subsection{Data Sets Description}
To evaluate the effectiveness of the BPDG method, extensive experiments are conducted based on the PersonalDialog dataset\cite{Zheng}. This corpus contains sparse personas of multi-party, where the personalized responses in dyadic dialogues involve bilateral personalized information. It is very challenging to choose which persona to generate, so we pick dyadic dialogues from the original corpus for our research. This dataset provides personalized profiles of both speakers, including three personal attributes i.e., ``Gender", ``Area" and ``Individual interests".
\par Since in some cases of the original corpus, the personalized profiles are missing. We construct a heuristic script to select the data with complete personalized information of both parties. The constructed dialogue dataset is referred to as the bilateral personalized dataset in this article. The bilateral personalized dataset consists of 410K dialogues in total, where 400K is randomly sampled as the training set, and the rest 10K data as the validation set. The average length of each dialogue is about 3.5 rounds and the average length of each sentence is about 7.45 characters.
\par The evaluation settings of the ECDT\footnote{http://conference.cipsc.org.cn/smp2019/evaluation.html} are adopted, to test the performance of different methods in different contexts. Specifically, two test sets\footnote{https://worksheets.codalab.org/worksheets/0x8f68b61a8b2249d7b314c6e8 \ 00e2dace} (i.e., a random set and a biased set) are constructed for the evaluation. The random set is a collection of dialogues between both parties, most of which do not contain personas. It is constructed for testing the performance of different methods in a context where the two interacting parties do not intentionally show their personas. The biased test set is the dialogue set between both parties with personalized information, where the speaker tends to reveal personalized information about both parties during the conversation. It is used for testing the performance of different methods in the context where the speakers intentionally show their personas.
\subsection{Bilateral Persona Classifier}
To better evaluate whether the response is bilateral persona-consistent or not, we design the bilateral persona classifier $P_{\pi}$ as an objective metric, which is trained with the aforementioned personalized labels. Each sentence is labeled with one of the three labels: 0 for the sentence related to the persona of the user, 1 for the sentence related to the persona of the robot, and 2 for the sentence that does not contain the persona.
\par The bilateral persona classifier is used to evaluate whether the response $Y$ contains the user persona $U$ or the robot persona $R$. To calculate each probability of the respective category, the response Y with bilateral personas is concatenated with $< SEP >$. 
After calculating each probability, the probability of category 0 and category 1 is added together as the probability of the bilateral personalized response. 
About 10K rounds of dialogues containing bilateral personas are randomly sampled from the bilateral personalized dataset, where the category ratio is 1:1:3. Then, we divide the above data into training, validation, and test set at a ratio of 8:1:1 to train the bilateral persona classifier. The accuracy of the classifier on the test set reaches $90.2\%$ in a 5-fold cross-validation setting.
\subsection{Content Relevance Classifier}
The content relevance classifier is used for ranking the candidates under the criterion of the CMIM. After the candidate list is generated by the BPDG model, we calculate the content relevance probability of each generated response co-occurring in the current dialogues under the conditions of the personalized history $H$, the user persona $U$, and the robot persona $R$. These conditions and each generated response are concatenated with $<SEP>$ for calculating the content relevance probability. After the probability of each generated response is calculated, the final response is selected to output. Specifically, the content relevance classifier is trained on the bilateral personalized dataset, using the ERNIE-base model \cite {38} to finetune in the labeled dialogues. In the 5-fold cross-validation setting, the accuracy reaches $80.4\%$.
\subsection{Implementation Details}
We implement all the experiments of the bilateral personalized dialogue generation with the pre-train model called LCCC-base \cite{LCCC}, which a Chinese pre-trained model based on the GPT2 framework with a vocab of 13088 characters, is used to initialize the parameters of the encoder and decoder with  transfer learning. According to \cite{30}, the shared weights of the encoder and decoder are adopted in this article, as it is beneficial for improving the quality of generated responses. The encoder and decoder include 12 Transformer blocks, among which the self-attention heads are 12. The size of the token embedding is 768 and the context window is 512. The parameter ${d_{model}} = 512$, ${n} = 64$, $\lambda_{1}$ = 0.2, $\lambda_{2}$ = 0.5, and $\lambda_{3}$ = 0.3. The diverse beam search strategy adopted in the proposed method is to generate the candidate list with the BPDG model, where the beam size is set to 20 and the group size is set to 4. The content relevance classifier is to calculate the relevance probability for each sentence in the candidates list under the criterion of the CMIM. The final generated response $Y^{*}$ is selected to output. The BPDG model is finetuned directly on the bilateral personalized dataset for 30 epochs, where the batch size is 64 with gradient accumulation, using the Noam optimization scheduler\cite{39} with 2000 warm-up steps on two NVIDIA 2080 Ti GPUs. All the experimental codes are released at
\href{https://github.com/Lireanstar/BPDG}{https://github.com/Lireanstar/BPDG}\footnote{Code and data will be publicly available}.
\subsection{Compared Methods}
Several state-of-the-art baseline methods are compared with ours. These methods are described below:
\par
\begin{enumerate}
	\item  \emph{S2S + Atten.}: This method applies a three-layer Bi-GRU to project the input text into embeddings with a fixed size. Another three-layer GRU utilizes an attention mechanism\cite{40} for response generation. The word embedding parameters of encoder and decoder are initialized by the pre-trained word vector\footnote{https://github.com/Embedding/Chinese-Word-Vectors}. 
	The parameter weights of the GRU network are initialized with a uniform distribution [-0.05, 0.05]. The model is optimized by implementing the Adam optimization scheduler.
	\item \emph{Trans.}: The Trans. employs the Transformer \cite{31} using the self-attention mechanism to generate responses. The model is initialized with the uniform distribution [-0.02, 0.02] and takes the concatenated dialogue history as input without using personas. We optimize the model by implementing the Noam\cite{39} optimization scheduler.
	\item \emph{TTransfo.}: The TTransfo. is introduced by \cite{25} optimizing a multi-task object for training. This model is initialized by the LCCC-base pre-trained model and finetunes directly on the bilateral personalized dataset only with the concatenated history. The Norm optimization scheduler is used for  training the model with gradient accumulation (with batch size 64).
	\item \emph{LConv.}: The LConv. represents the multi-input model proposed in \cite{24}. This model is initialized with the LCCC-base pre-trained model, which shares the parameters of the encoder and decoder. The model finetunes directly on the bilateral personalized dataset with the concatenated dialogue history. The Norm optimization scheduler is used for training the model with gradient accumulation (with batch size 64).
	\item \emph{TTransfo.+P}: It extends the TTransfo. by incorporating the speaker's persona. When fine-tuning, the contextual dialogues concatenated with the speaker's personalized information are input into the model. The Norm optimization scheduler is implemented for training, where the batch size is set to 64 with gradient accumulation.
	\item \emph{LConv.+P}: It extends the LConv. by incorporating the speaker's persona. When fine-tuning the contextual dialogues concatenated with the speaker's personalized information are input into the model. The Norm optimization scheduler is implemented for training, where the batch size is set to 64 with gradient accumulation.
	\item \emph{PWDWP}: The PWDWP\cite{ZhengAAAI} is initialized by the LCCC-base pre-trained model and finetunes on the bilateral personalized dataset. This model incorporates personalized attributes embedding in the dialogue context for each speaker and devises a persona routing to weigh the persona-related encodings that are input into the decoder. The Norm optimization scheduler is implemented for training, where the batch size is set to 64 with gradient accumulation. This model is similar to our method, which is the strong baseline method in the explicit personalized dialogue system.
\end{enumerate}
\section{Automatic Evaluation}
In order to fully evaluate the effectiveness of the proposed method compared with the baseline methods, we choose various metrics for the automatic evaluation. In this section, we introduce these metrics and give a detailed analysis of the results.
\subsection{Objective Metrics Introduction}
\par \emph{1) Bi-Persona Acc}

\par The Bi-Persona Acc (BPAcc) is used to measure the degree of personalization in the response. We extend the unilateral persona-consistent \cite{ZhengAAAI}, which represents that the persona is consistent with the speaker, to the bilateral persona-consistent. The Bi-Persona Acc represents the bilateral persona classification accuracy of the sentence which is not only consistent with the speaker's persona but also with the persona of the other party. Each generated response and the bilateral personas are input into the bilateral persona classifier to obtain the Bi-Persona Acc. Therefore, we add the user and robot persona classification accuracy together to obtain the possibility of the response that contains bilateral personalized information. The higher Bi-Persona Acc score means that the generated response is more personalized and more likely to be bilateral persona-consistent.
\begin{equation}
\text {BPAcc}= \frac{\text { $P_{\pi}$(1) +$ P_{\pi}$(2) }}{\text { $P_{\pi}(0)$ }+\text { $P_{\pi}$(1)}+\text { $P_{\pi}$(2)}}
\label{f1}
\end{equation}
where $P_{\pi}(x)$ represents the output probability with label $x$ of bilateral persona classifier.
\par \emph{2) BLEU}
\par The BLEU (bilingual evaluation understudy) \cite{41} is utilized to evaluate the quality of the text in translation. In dialogue generation, the BLEU is calculated with the weighted n-gram overlap between the ground-truth response $\widehat{Y}$ and generated responses $Y^{*}$. The n-gram calculation is shown in the equation (\ref{bleu}):
\begin{equation}
P_{n}(\widehat{Y}, Y^{*})=\frac{\sum_{k} \min \left(\text {Cnt}_{\text {clip}}(k,\widehat{Y} ), \text {Cnt}_{\text {clip}}(k, Y^{*})\right)}{\sum_{k} \text {Cnt}(k, \widehat{Y})}
\label{bleu}
\end{equation}
where $k$ traverses all the n-grams candidates, the $\widehat{Y}$ and the $Y^{*}$ represent the ground-truth response and the generated response respectively, $\text {Cnt}_{\text {clip}}(k, Y^{*})$ represents the clipped n-grams number in the generated response $Y^{*}$, $\text {Cnt}(k, \widehat{Y})$ represents n-grams number in the ground-truth response $\widehat{Y}$. The weight $BP(\widehat{Y}, Y^{*})$ can be calculated as equation (\ref{bp}):
\begin{equation}
BP(\widehat{Y}, Y^{*})=\begin{cases}
1\text{,}& \text { if }|Y^{*}|>|\widehat{Y}| \\
e^{(1-|\widehat{Y}| / \mid Y^{*} \mid}\text{,} & \text { if }|Y^{*}| \leq|\widehat{Y}|
\end{cases}
\label{bp}
\end{equation}
where $|Y^{*}|$ represents the length of the generated response, $|\widehat{Y}|$ represents the length of the ground-truth response. The BLEU is calculated as follows:
\begin{equation}
BLEU=B P(\widehat{Y}, Y^{*}) \cdot \exp \left(\sum_{n=1}^{N} w_{n} \log P_{n}(\widehat{Y}, Y^{*})\right)
\end{equation}
where $N$ is set to 2 and the weighted factor $ w_{n}$ is set to $1/N$, the percentile fraction we use is set to 1000, which is the same settings as the NLTK\footnote{https://github.com/nltk/nltk}. The higher the BLEU score, the better the quality of the generated response.
\par \emph{3) F1}
\par The F1 score is implemented to measure the accuracy of the model on the data set compared to the ground truth, which includes two parts: precision and recall. The precision means the proportion of words in the generated response contained in the ground-truth response, and the recall means the proportion of words in the ground-truth response contained in the generated response. The calculation of F1 score is the same as \cite{42} and can be written as equation (\ref{f1}):
\begin{equation}
\text {F1}=2 \times \frac{\text { precision $\cdot$ recall }}{\text { precision }+\text { recall }}
\label{f1}
\end{equation}
\par \emph{4) Distinct}
\par The Distinct \cite{35} is adopted to measure the average score of the sum of unique unigrams and bigrams contained in the generated responses, which is divided by the total number of generated words. The equation can be written as follows:
\begin{equation}
\text { Distinct }=\frac{1}{2}\times\frac{Cnt(U_{uni}) +Cnt(U_{bi})}{Num_{tokens}}
\end{equation}
where the $Cnt(U_{uni})$ represents the number of unigrams that are not repeated in the response compared with the ground-truth response, the $Num_{tokens}$ represents the total number of generated words, the higher the distinct score, the more specific and diverse the response generated.
\par \emph{5) PPL}
\par The PPL (perplexity) \cite{huang2020cotk} is widely used to measure the performance that the model predicts different utterances in the test set. For the ground-truth response $\widehat{Y}=\left\{y_{1}, y_{2}, \ldots, y_{m}\right\}$, the perplexity is calculated by the trained model, and can be calculated as equation (\ref{ppl}):
\begin{equation}
\begin{array}{c}
\text { Perplexity }=\exp \left(-\frac{1}{N} \sum_{i=1}^{m} t_{i}\right)
\end{array}
\label{ppl}
\end{equation}
\begin{equation}
t_{i}=\begin{cases}
\log P\left(y_{i}\right)+\varepsilon\text{,} & \text { if } y_{i} \in F \\
\log (P(unk) /|R|)+\varepsilon\text{,} & \text { if } y_{i} \in R
\end{cases}
\end{equation}
where the $F$ represents the set of words in the frequent vocabulary and the $R$ represents the set of words that are in the rare vocabulary, $P(unk)$ represents the logits of unknown token predicted by the model. $|R|$ is the number of words that are in the rare vocabulary, the $\varepsilon$ is set to $10^{-8}$, which is used to ensure that logits are not zero.
\subsection{Results and Analysis}
\begin{table}
	\centering
	\renewcommand\arraystretch{1.2}
	\caption{Experimental Results of Five Metrics on \protect \\
		Random Test Set}
	\begin{tabular}{cccccc}
		\specialrule{0.1em}{1pt}{1pt}
		Method        & BPAcc          & BLEU           & F1             & Distinct          & PPL           \\
		\specialrule{0.05em}{0.5pt}{0.5pt}
		S2S + Atten.  & 9.71           & 3.18           & 4.32           & 0.098          & 87.41          \\
		Trans.        & 11.84          & 3.27           & 7.48           & 0.187          & 77.55          \\
		TTransfo.    & 15.43          & 6.33           & 10.40          & 0.255          & \textbf{47.48} \\
		TTransfo.+P & 44.68          & 6.27           & 10.11          & 0.251          & 52.14          \\
		LConv.        & 12.41          & 3.38           & 7.52           & 0.243          & 51.32          \\
		LConv.+P    & 51.36          & 6.02           & 8.81           & 0.234          & 55.04          \\
		PWDWP        & 63.13          & 10.56          & 10.47          & 0.280          & 60.33          \\
		\specialrule{0.05em}{0.5pt}{0.5pt}
		Ours         & 64.63          & \textbf{11.66} & \textbf{11.30} & \textbf{0.282} & 61.93          \\
		Ours, $\alpha$=1    & \textbf{87.12} & 10.64          & 10.97          & 0.279          & 60.55          \\
		Ours, $\beta$=1    & 82.42          & 10.32          & 10.52          & 0.274          & 60.12          \\
		Ours, $\gamma$=1    & 52.11          & 8.34           & 10.22          & 0.272          & 59.92         \\
		\specialrule{0.1em}{1pt}{1pt}
	\end{tabular}
	\label{tab1}
\end{table}
\begin{table}
	\centering
	\caption{Experimental Results of Five Metrics on \protect \\
		Baised Test Set}
	\renewcommand\arraystretch{1.2}
	\begin{tabular}{cccccc}
		\specialrule{0.1em}{1pt}{1pt}
		Method       & BPAcc          & BLEU           & F1             & Distinct          & PPL           \\
		\specialrule{0.05em}{0.5pt}{0.5pt}
		S2S + Atten. & 22.42          & 4.32           & 4.71           & 0.116          & 97.53          \\
		Trans.       & 29.14          & 6.89           & 8.07           & 0.251          & 79.23          \\
		TTransfo.   & 43.12          & 15.22          & 15.93          & 0.268          & \textbf{49.59} \\
		TTransfo.+P  & 60.25          & 15.55          & 16.27          & 0.260          & 57.63          \\
		LConv.       & 42.73          & 10.03          & 9.09           & 0.270          & 56.41          \\
		LConv.+P     & 57.43          & 12.13          & 10.02          & 0.263          & 58.74          \\
		PWDWP       & 88.19          & 17.21          & 16.40          & 0.285          & 60.81          \\
		\specialrule{0.05em}{0.5pt}{0.5pt}
		Ours        & 92.14          & 24.64          & 18.05          & \textbf{0.287} & 58.59          \\
		Ours, $\alpha$=1   & \textbf{93.75} & 23.51          & 17.97          & 0.286          & 61.77          \\
		Ours, $\beta$=1   & 91.12          & \textbf{24.72} & \textbf{18.52} & 0.285          & 60.11          \\
		Ours, $\gamma$=1   & 57.62          & 17.51          & 17.01          & 0.282          & 62.66   \\
		\specialrule{0.1em}{0.5pt}{1pt}
	\end{tabular}
	\label{tab2}
\end{table}
\par Table \ref{tab1}. and Table \ref{tab2}. respectively show the comparison results of the proposed method and different baseline methods on five metrics, and also present the performance of our method with different persona-aware weights. It can be seen from the results that, compared with the baseline methods, our method is superior to all metrics except the PPL. Noted that the ppl. score is inconsistent with \cite{ZhengAAAI}, because they have used external personalized corpus for pre-training, while this pre-training corpus is not open source. Table \ref{tab1}. and Table \ref{tab2}. show that we have used the open-source LCCC model for initializing all the baseline models.
\par Under the same experimental conditions, further conclusions are that: (1) under the same automatic weighting setting, our method is better than the strong baseline method (i.e., PWDWP). On the random set, it outperforms with $1.5\%$ in BPAcc, $1.1\%$ in BLEU, $0.83\%$ in F1, and $0.2\%$ in Distinct. While on the biased set, our method outperforms with $3.95\%$ in BPAcc, $7.43\%$ in BLEU, $1.65\%$ in F1, and $0.2\%$ in Distinct. Especially on the baised set, our method is superior to the compared baseline methods. This shows that our method can generate more personalized and better responses. 
(2) It can be found that both in Table \ref{tab1}. and Table \ref{tab2}. the PPL scores in bold (i.e., 47.48 and 49.59) show that the best results of the PPL  appear on the TTransfo, which is the method without incorporating the personalized information. However, the methods with personalized information (i.e., TTransfo.+P, LConv.+P, PWDWP, and our method) all obtain the higher PPL score. This indicates that generating responses with personalized information will hurt the PPL score. It occurs because the words involving the persona in social conversation are relatively rare. Such words may bring bias and lead to the worse perplexity score, which is in line with the results in \cite{ZhengAAAI, 42}. The baseline methods with a lower perplexity score tend to generate more general responses, thus they cannot generate responses that match the bilateral personas. As a result, the BPAcc scores of these baseline methods are relatively low.
%
(3) Compared with the methods without personalized information (i.e., S2S + Atten., TTransfo. and LConv.), the methods with unilateral personalized information  (i.e., TTransfo.+P, LConv.+P, and PWDWP) on the two test sets get higher BPAcc scores. Moreover, the method with bilateral personalized information (i.e., our method) has a higher BPAcc score on the two test sets than the strong baseline method with unilateral personalized information (i.e., PWDWP). 
This indicates the effectiveness of the proposed bilateral persona classifier to evaluate the degree of personalization and bilateral-consistent.
(4) On the random set, the proposed method outperforms the other baseline methods that only incorporate the unilateral persona in BPAcc (i.e., 87.12 in bold). Similar trends are observed on the biased set (i.e., 93.75 in bold), which indicates that incorporating the other party's personalized information in the decoding process is beneficial to generate more personalized and more bilateral persona-consistent responses.
(5) The proposed different persona-aware weights (i.e., $\alpha$, $\beta$, and $\gamma$) can be used to control the persona presented in the generated response. The results of the two test sets show that under different context settings, it will improve the effect of personalized response generation with different persona-aware weights. This indicates that the proposed dynamic persona-aware fusion module is beneficial to generate diversified dialogue responses rich in bilateral personalized information.
\subsection{Ablation Study}
In order to test the performance of different modules on the proposed method, several ablation experiments are implemented as follows.
(1)	Each module of multi-task settings is deleted respectively, including the language model (w/o LM) and the dynamic persona-aware fusion module (w/o PAF).
(2)	The pre-trained model is also deleted (w/o PreT) to test the performance of transfer learning. 
(3)	The dialogue utterance with corresponded personas embedding (w/o PEmb) and the conditional mutual information maximum criterion (w/o CMIM) are deleted respectively to test the effect of different strategies on the BPDG method.
\begin{table}
	\centering
	\renewcommand\arraystretch{1.2}
	\caption{Ablation Results of Our Proposed Method \protect \\
		on Random Test Set}
	\begin{tabular}{cccccc}
		\specialrule{0.1em}{1pt}{1pt}
		Method               & BPAcc                 & BLEU                 & F1                   & Distinct                & PPL                 \\
		\specialrule{0.05em}{0.5pt}{0.5pt}
		BPDG                  & 64.63                & 11.66                & 11.30                & 0.282                & 61.93                \\
		w/o LM               & 40.42                & 9.21                 & 10.27                & 0.273                & 60.17                \\
		w/o PAF            & 47.12                & 8.06                 & 10.51                & 0.267                & 57.34                \\
		w/o PreT               & 42.19                & 4.71                 & 8.74                 & 0.247                & 88.89                \\
		w/o PEmb             & 43.24                & 9.12                 & 11.01                & 0.270                & 63.12                \\
		w/o CMIM             & 60.14                & 8.13                 & 8.83                 & \textbf{0.301}       & 61.93            \\
		\specialrule{0.1em}{1pt}{1pt}
	\end{tabular}
	\label{tab3}
\end{table}
\begin{table}
	\centering
	\renewcommand\arraystretch{1.2}
	\caption{Ablation Results of Our Proposed Method \protect \\
		on Biased Test Set}
	\begin{tabular}{cccccc}
		\specialrule{0.1em}{1pt}{1pt}
		Method    & BPAcc   & BLEU  & F1    & Distinct          & PPL   \\
		\specialrule{0.05em}{0.5pt}{0.5pt}
		BPDG       & 92.14 & 24.64 & 18.05 & 0.287          & 58.59  \\
		w/o LM    & 69.54 & 20.96 & 17.87 & 0.254          & 62.50  \\
		w/o PAF & 77.22 & 16.95 & 17.15 & 0.240          & 60.34  \\
		w/o PreT    & 74.57 & 13.56 & 15.12 & 0.232          & 77.52  \\
		w/o PEmb  & 74.41 & 18.79 & 17.44 & 0.249          & 61.19  \\
		w/o CMIM  & 87.11 & 17.22 & 16.12 & \textbf{0.312} & 58.59
		\\
		\specialrule{0.1em}{1pt}{1pt}
	\end{tabular}
	\label{tab4}
\end{table}
\par Table \ref{tab3}. and Table \ref{tab4}. show the ablation results.
From the results, the further conclusion can be drawn that: (1) the LM module learns the language's semantics from the dialogue context. Without the LM module, it will hurt the dynamic persona-aware fusion on the BPDG method. As a result, the BPAcc score will be decreased most.
(2) The PAF module is beneficial to generate more personalized and diversified responses. The above different modules of multi-task learning prove to improve the total effect of personalized dialogue generation.
(3) The pre-trained language model provides a good parameter initialization for the BPDG method, which helps to improve training efficiency by transferring the knowledge of the original domain to the target domain.
(4) The PEmb strategy improves the final performance by embedding the personalized attributes to the corresponded dialogue utterances. 
(5) More importantly, the CMIM criterion is effective to improve the BPAcc, BLEU, and F1 scores, but it may decrease the Distinct scores, which are bolded in Table \ref{tab3}. and Table \ref{tab4}. This is because the sorting and selection steps from the candidates may hurt the diversity of the generated responses.
%
%
%
\begin{figure*}[ht]
	\centering{
		{\includegraphics[width=14cm]{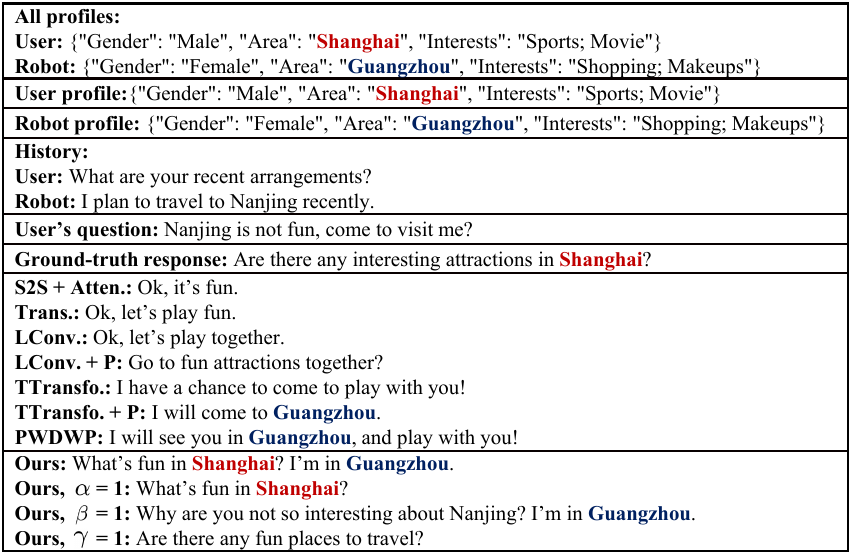}}
	}
	\caption{Sampled responses generated by baseline methods and our method.}
	\label{fig6}
\end{figure*}
\section{Human Evaluation}
We also perform a human evaluation to test the quality of responses generated by different methods. In this section,
we introduce these metrics and give a comprehensive analysis of the results.
\begin{table}[]
	\renewcommand\arraystretch{1.2}
	\caption{Human Evaluation on the Random and Biased Test Set. \protect \\ With * Indicate the Significant Difference With
		\protect \\ the Best Result (t-test and p-value $<$ 0.05)}
	\centering
	\begin{tabular}{lp{0.6cm}lp{0.6cm}lp{0.6cm}lp{0.6cm}lp{0.6cm}lp{0.6cm}lp{0.6cm}}
		\specialrule{0.1em}{1pt}{1pt}
		\multicolumn{1}{c}{\multirow{3}{*}{Method}} & \multicolumn{2}{c}{Sentence}                        & \multicolumn{2}{c}{Bilateral Persona}               & \multicolumn{2}{c}{Context}                         \\
		\multicolumn{1}{c}{}                       & \multicolumn{2}{c}{Fluency}                         & \multicolumn{2}{c}{Consistency}                      & \multicolumn{2}{c}{Consistency}                       \\ \cline{2-7} 
		\multicolumn{1}{l}{}                       & \multicolumn{1}{c}{Rand} & \multicolumn{1}{c}{Biasd} & \multicolumn{1}{c}{Rand} & \multicolumn{1}{c}{\ Biasd} & \multicolumn{1}{c}{Rand} & \multicolumn{1}{c}{Biasd} \\ 
		\hline
		S2S + Atten.                                 & 1.24*          & 1.14*          & 0.72*             & \  \ 0.91*            & 0.95*          & 1.12*          \\
		Trans.                                      & 1.39*          & 1.42*         &  0.88*              & \ \ 1.07*         & 1.12*          & 1.38*          \\
		TTransfo.                                    & 1.41*           & 1.37*         & 0.92*              &\ \ 1.19*           & 1.04*          & 1.41          \\
		TTransfo.+P                                   & 1.42*          & 1.43*          & 0.96*              & \ \ 1.27*             & 1.05*             & 1.39*          \\
		LConv.                                     & 1.55*          & 1.54*          & 1.08*              & \ \ 1.42          & 1.01*          & 1.58          \\
		LConv.+P                                      & 1.70*           & 1.66*          & 1.12*              &  \ \ 1.45              & 1.23          & 1.60*           \\
		PWDWP                                      & 1.79          & 1.82         & 1.35              &   \ \  1.63*             & 1.25*          & 1.71          \\ 			
		\specialrule{0.05em}{0.5pt}{0.5pt} 
		Ours                                      & \textbf{1.82} & 1.87*         & 1.41               &   \ \  1.74              & \textbf{1.27} & \textbf{1.75} \\
		Ours, $\alpha$=1                                         & 1.79*          & \textbf{1.89}* & 1.42              &  \ \  \textbf{1.80}*    & 1.18*          & 1.71*           \\
		Ours, $\beta$=1                                       & 1.81          & 1.85         & \textbf{1.47}     &    \ \ 1.72*              & 1.16*          & 1.67          \\ 
		Ours, $\gamma$=1                                   & 1.76*          & 1.78*          & 1.09*              &    \ \ 1.32*              & 1.24          & 1.69          \\ 
		\specialrule{0.05em}{0.5pt}{0.5pt}
		Human Resp                                         & 1.89          & 1.90         &  1.07*              &    \ \ 1.78             & 1.67          & 1.88    \\
		\specialrule{0.1em}{0.5pt}{1pt}
	\end{tabular}
	\label{tab5}
\end{table}
\subsection{Subjective Metrics Introduction}
The evaluation metrics we choose are from three aspects, as is shown below.
\par \emph{1) Sentence fluency}
\par Sentence fluency represents the fluency of responses generated by different methods.
\par \emph{2) Bilateral persona consistency}
\par Bilateral persona consistency indicates whether the information is consistent with the user's or the robot's personalized information when generating a response by different methods.
\par \emph{3) Context consistency}
\par Context consistency means whether the response generated by different methods is consistent with the dialogue context.
\par Three annotators are required to rate the quality of the responses according to the following three rating criteria: (1) +2: the response is not only semantically and grammatically related, but also bilateral persona-consistent. (2) +1: the response satisfies the grammatical rules and can be used as a response, but is too general and trivial. (3) +0: the response is semantically irrelevant, ungrammatical, or conflicts with the personalized information.

\subsection{Results and Analysis}

%

We sample 100 dialogue sessions from the original random and biased test set respectively for the human evaluation. The inter-annotator agreement is measured with Fleiss's kappa $\kappa$ \cite{43}. Particularly, the $\kappa$ value for sentence fluency, bilateral persona consistency, and context consistency is 0.81, 0.71, 0.64 on the random test set respectively, and 0.75, 0.67, 0.61 on the biased test set respectively. The results indicate that the sentence fluency, the bilateral persona consistency, and the context coherency of two test sets achieve substantial annotation agreement.
\par Table \ref{tab5}. shows the results of the human evaluation that the proposed method outperforms all baseline methods in all human metrics (t-test and p-value $<$ 0.05).
Further observations indicate that (1) incorporating bilateral personas into the generated response will impair the sentence fluency and the context consistency, which corresponds to the high BPAcc score and the low PPL score in the automatic evaluation. Despite this, our method has achieved significant advantages in fluency and context consistency in two test sets compared with other methods.
(2) The proposed dynamic persona-aware fusion module is designed to control different persona-aware weights
for the personalized response generation. This module contributes to better bilateral persona consistency.
At the same time, the bilateral persona consistency outperforms the human in the random test set and the test set. This shows that the proposed dynamic persona-aware fusion module is conducive to generating more personalized responses in both dialogue contexts. This observation is also in line with the BPAcc in automatic evaluation shown in Table \ref{tab1}. and Table \ref{tab2}. (3) Compared with the PWDWP method, the proposed BPDG has a great improvement in context consistency. This is due to the effect of the CMIM criterion, which selects the response from the generated the candidate list under the condition of the bilateral personas and the context. This observation also corresponds with the automatic evaluation results of BLEU and F1 metrics shown in Table \ref{tab3}. and Table \ref{tab4}.

\subsection{Case Study}
The case study is shown in Fig. \ref{fig6}. The proposed method can generate a response consistent with the personas of both parties in the conversation.
As we can see, the response generated by the \text{TTransfo.+P} and the PWDWP methods may be unilateral persona-consistent without incorporating the persona of the other party. The other baseline methods (i.e., S2S + Atten., TTrans., TTransfo., LConv., LConv.+P) may also generate a general response that lacks personalized information. The proposed BPDG method utilizes bilateral personalized information to generate responses that are in line with human cognition while constraining the contents of the generated responses with the CMIM criterion. Specifically, given the user input and the bilateral personas, our method can control the generated response content with different persona-aware weights. The $\alpha =1$ means that the user's personalized information presents in the response, such as Shanghai. The $\beta=1$ means that the robot's personalized information presents in the response such as Guangzhou. The $\gamma = 1$ means that the personalized information does not present in the response, but it is relevant to the context, such as travel.
\section{Conclusion}
In this article, we propose the BPDG method to generate more personalized and bilateral persona-consistent responses. Our method utilizes transfer learning via initializing parameters with a pre-trained model, then jointly trains the encoder, the dynamic persona-aware fusion module, and the decoder with multi-task learning. Experiments show that the transfer learning and multi-task learning method are conducive to improving the performance of dialogue in various metrics. In addition, the generated candidate responses are selected with the CMIM criterion, which shows that the quality of the final response can be greatly improved. Extensive experiments are conducted to measure the effectiveness of the BPDG method, which shows that the BPDG has advantages in all metics but the PPL. The human evaluation results also prove that the BPDG method generates more fluent, context-consistent, and bilateral persona-consistent responses than several state-of-the-art methods.
\par It is worth noting that in open-domain dialogue, the human response is one-to-many, and the open-domain corpus cannot contain all the situations. Moreover, people will respond and reason based on existing information during the conversation. In the future, we will explore other fusion strategy based dialogue generation methods with comprehensive reasoning of the existing information, to further improve the quality of the generated response.
{
	\bibliographystyle{IEEEtran}
	\bibliography{bipbib}
}
\end{document}